\documentclass[letterpaper]{article}
\renewcommand{\subparagraph}{\paragraph}
\usepackage{theapa}
\usepackage{times}
\usepackage{amsmath, amsthm, amssymb}
\usepackage{pstricks}
\usepackage{pst-tree}
\usepackage{bm}
\usepackage{graphicx}
\usepackage{epsfig}
\usepackage{subfigure}
\usepackage{amsmath}
\usepackage{amsfonts}
\usepackage{color}
\usepackage{array}
\usepackage{colortbl}
\usepackage{framed}
\usepackage{url}
\usepackage{booktabs}
\usepackage{multirow}
\usepackage{dsfont}

\usepackage{multicol}
\usepackage{lscape}
\usepackage{relsize}

\usepackage{nicefrac}
\newcommand{\vect}[1]{\underline{\smash{#1}}}
\renewcommand{\v}[1]{\vect{#1}}
\newcommand{\reals}{\mathds{R}}
\newcommand{\sX}{\mathcal{X}}
\newcommand{\sD}{\mathcal{D}}
\newcommand{\br}{^{\text{\textnormal{ r}}}}
\newcommand{\cat}{^{\text{\textnormal{c}}}}
\newcommand{\cut}[1]{}
\newcommand{\hide}[1]{}

\usepackage{rotating}

\DeclareFontFamily{OT1}{pzc}{}
\DeclareFontShape{OT1}{pzc}{m}{it}{<-> s * [1.200] pzcmi7t}{}
\DeclareMathAlphabet{\mathscr}{OT1}{pzc}{m}{it}

\newcommand\transpose{{\textrm{\tiny{\sf{T}}}}}
\newcommand{\note}[1]{}

\renewcommand{\note}[1]{~\\\frame{\begin{minipage}[c]{\textwidth}\vspace{2pt}\center{#1}\vspace{2pt}\end{minipage}}\vspace{3pt}\\}

\usepackage{amsmath, amsthm, amssymb}
\newtheorem{thm}{Theorem}%[section]

\newtheorem{prop}[thm]{Proposition}

\newtheorem{define}[thm]{Definition}
\newcommand{\indicator}{\ensuremath\mathds{I}}

\begin{document}

\title{A Kernel for Hierarchical Parameter Spaces}

\author{Frank Hutter and Michael A. Osborne\\
{\tt fh@informatik.uni-freiburg.de} and {\tt mosb@robots.ox.ac.uk}
}

\date{September 7, 2013}

\maketitle
\begin{abstract}
\noindent{}We define a family of kernels for mixed continuous/discrete hierarchical parameter spaces and show that they are positive definite.
\end{abstract}

%%%%%%%%%%%%%%%%%%%%%%%%%%%%%%%%%%%%%%%%%%%%%%%%%%%%%%%%%%%%%%%%%%%%%%%%%%%%%%%%%%%%%%%%%%%%%%%%%%%%%%%%%%%%%%%%%%%%%%%%%%%%%
\section{Introduction}
%%%%%%%%%%%%%%%%%%%%%%%%%%%%%%%%%%%%%%%%%%%%%%%%%%%%%%%%%%%%%%%%%%%%%%%%%%%%%%%%%%%%%%%%%%%%%%%%%%%%%%%%%%%%%%%%%%%%%%%%%%%%%

We aim to do inference about some function $g$ with domain (input space) $\sX$. $\sX = \prod_{i=1}^D \sX_i$ is a $D$-dimensional input space, where each individual dimension is either bounded real or categorical, that is, $\sX_i$ is either $[l_i, u_i] \subset \reals$ (with lower and upper bounds $l_i$ and $u_i$, respectively) or $\{v_{i,1}, \dots, v_{i,m_i}\}$. 

Associated with $\sX$, there is a DAG structure $\sD$, whose vertices are the dimensions $\{1,\,\ldots,\,D\}$. $\sX$ will be restricted by $\sD$: if vertex $i$ has children under $\sD$, $\sX_i$ must be categorical. $\sD$ is also used to specify when each input is \emph{active} (that is, relevant to inference about $g$). In particular, we assume each input dimension is only active under some instantiations of its ancestor dimensions in $\sD$. More precisely, we define $D$ functions $\delta_i\colon \sX\to \mathcal{B}$, for $i \in \{1,\,\ldots,\,D\}$, and where $\mathcal{B} = \{\text{true}, \text{false}\}$. We take 
\begin{equation}
 \delta_i(\v{x}) = \delta_i\bigl(\v{x}(\text{anc}_i)\bigr),
\end{equation}
where $\text{anc}_i$ are the ancestor vertices of $i$ in $\sD$, such that $\delta_i(\v{x})$ is true only for appropriate values of those entries of $\v{x}$ corresponding to ancestors of $i$ in $\sD$. We say $i$ is active for $\v{x}$ iff $\delta_i(\v{x})$.

%if all its parent dimensions $P_i$ are active themselves and each parent $p\in P_i$ takes one of the values in the finite set $V_{i,p}$. 
Our aim is to specify a kernel for $\sX$, \emph{i.e.}, a positive semi-definite function  $k\colon \sX \times \sX \to \reals$. We will first specify an individual kernel for each input dimension, \emph{i.e.}, a positive semi-definite function $k_i\colon \sX \times \sX \to \reals$. $k$ can then be taken as either a sum,
\begin{equation}
 k(\v{x}, \v{x}') = \sum_{i=1}^D k_i(\v{x},\v{x}'),
\end{equation}
product,
\begin{equation}
 k(\v{x}, \v{x}') = \prod_{i=1}^D k_i(\v{x},\v{x}'),
\end{equation}
or any other permitted combination, of these individual kernels. Note that each individual kernel $k_i$ will depend on an input vector $\v{x}$ only through dependence on $x_i$ and $\delta_i(\v{x})$,
\begin{equation}
  k_i(\v{x},\v{x}') = \tilde{k}_i\bigl(x_i,\delta_i(\v{x}),x_i', \delta_i(\v{x}') \bigr).
\end{equation}
That is, $x_j$ for $j\neq i$ will influence $k_i(\v{x},\v{x}')$ only if $j \in \text{anc}_i$, and only by affecting whether $i$ is active.

Below we will construct pseudometrics $d{_i}\colon \sX \times \sX \to \reals^+$: that is, $d_i$ satisfies the requirements of a metric aside from the identity of indiscernibles. As for $k_i$, these pseudometrics will depend on an input vector $\v{x}$ only through dependence on both $x_i$ and $\delta_i(\v{x})$. $d{_i}(\v{x}, \v{x}')$ will be designed to provide an intuitive measure of how different $g(\v{x})$ is from $g(\v{x}')$. 
For each $i$, we will then construct a (pseudo-)isometry $f_i$ from
$\sX$ 
to a Euclidean space ($\reals^2$ for bounded real parameters, and $\reals^m$ for categorical-valued parameters with $m$ choices). That is, denoting the Euclidean metric on the appropriate space as $d{_E}$, $f_i$ will be such that
\begin{equation}
\label{eqn:d_i}
 d{_i}(\v{x},\v{x}')
=
d_{\text{E}}(f{_i}\bigl(\v{x}), f{_i}(\v{x}')\bigr)
\end{equation}
for all $\v{x}, \v{x}' \in \sX$. We can then use our transformed inputs, $f_i(\v{x})$, within any standard Euclidean kernel $\kappa$. We'll make this explicit in Proposition \ref{prop:psd_if_isometry}. 

\begin{define}
\label{def:psd_fun_euclid}
A function $\kappa\colon \reals^+ \to \reals$ is \emph{a positive semi-definite covariance function over Euclidean space} if $K \in \reals^{N\times N}$, defined by 
\begin{equation}
\nonumber K_{m, n} = \kappa\bigl(d_{\text{E}}(\v{y}_m, \v{y}_n)\bigr),\quad \text{for }
\v{y}_m, \v{y}_n \in \reals^P,\quad m, n = 1, \ldots, N, 
\end{equation}
is positive semi-definite for any $\v{y}_1, \dots, \v{y}_N \in \reals^P$. 
\end{define}

A popular example of such a $\kappa$ is the exponentiated quadratic, for which $\kappa(\delta) = \sigma^2 \exp(-\frac{1}{2} \frac{\delta^2}{\lambda^2})$; another popular choice is the rational quadratic, for which $\kappa(\delta) = \sigma^2 (1+\frac{1}{2\alpha} \frac{\delta^2}{\lambda^2})^{-\alpha}$.

\begin{prop}
Let $\kappa$ be a positive semi-definite covariance function over Euclidean space and let $d_i$ satisfy Equation \ref{eqn:d_i}. Then, 
$k_i\colon \sX \times \sX\to \reals^+$, defined by 
%\[k_i(\v{x},\v{x}') = \kappa( d_{\text{E}}(f_i(\v{x}), f_i(\v{x}')) )\]
\[k_i(\v{x},\v{x}') = \kappa\bigl( d_i(\v{x}, \v{x}') \bigr)\]
is a positive semi-definite covariance function over input space $\sX$. 
\label{prop:psd_if_isometry}
\begin{proof}
We need to show that for any $\v{x}_1, \dots, \v{x}_N \in \sX$, $K \in \reals^{N\times N}$ defined by
\begin{align*}
 K_{m, n} & = \kappa\bigl(d{_i}(\v{x}_m,\v{x}_n)\bigr)
,\quad \text{for }
\v{x}_m, \v{x}_n \in \sX,\quad m, n = 1, \ldots, N, 
\\
\intertext{is positive semi-definite. Now, by the definition of $d_i$,}
K_{m, n} & = \kappa\Bigl(d_{\text{E}}(f{_i}\bigl(\v{x}_m), f{_i}(\v{x}_n)\bigr)\Bigr) 
= \kappa\bigl(d_{\text{E}}(\v{y}_m, \v{y}_n)\bigr)
\end{align*}
where $\v{y}_m = f{_i}\bigl(\v{x}_m)$ and $\v{y}_n = f{_i}\bigl(\v{x}_n)$ are elements of $\reals^P$.
Then, by assumption that $\kappa$ is a positive semi-definite covariance function over Euclidean space, $K$ is positive semi-definite. 
\end{proof}
\end{prop}

We'll now define pseudometrics $d_i$ and associated isometries $f_i$ for both the bounded real and categorical cases.

%%%%%%%%%%%%%%%%%%%%%%%%%%%%%%%%%%%%%%%%%%%%%%%%%%%%%%%%%%%%%%%%%%%%%%%%%%%%%%%%%%%%%%%%%%%%%%%%%%%%%%%%%%%%%%%%%%%%%%%%%%%%%
\section{Bounded Real Dimensions}
%%%%%%%%%%%%%%%%%%%%%%%%%%%%%%%%%%%%%%%%%%%%%%%%%%%%%%%%%%%%%%%%%%%%%%%%%%%%%%%%%%%%%%%%%%%%%%%%%%%%%%%%%%%%%%%%%%%%%%%%%%%%%

Let's first focus on a bounded real input dimension $i$, i.e., $\sX_i=[l_i, u_i]$.
To emphasize that we're in this real case, we explicitly denote the pseudometric as $d\br_i$ and the (pseudo-)isometry from $(\sX, d_i)$ to $\reals^2,d_\text{E}$ 
as $f\br_i$. For the definitions, recall that $\delta_i(\v{x})$ is true iff dimension $i$ is active given the instantiation of $i$'s ancestors in $\v{x}$.

\begin{eqnarray}
\nonumber{}d\br_i(\v{x}, \v{x}') & = & \left\{\begin{array}{ll}
\nonumber{} 0 & \textrm{ if } \delta_i(\v{x}) = \delta_i(\v{x}') = \textrm{false}\\
\nonumber{} \omega_i & \textrm{ if } \delta_i(\v{x}) \neq \delta_i(\v{x}')\\
\nonumber{} \omega_i \sqrt{2} \sqrt{1 - \cos(\pi\rho_i \frac{x_i-x_i'}{u_i-l_i})} & \textrm{ if } \delta_i(\v{x}) = \delta_i(\v{x}') = \textrm{true}. \end{array}\right.
\end{eqnarray}

\begin{eqnarray}
\nonumber{}f_i\br(\v{x}) & = & \left\{\begin{array}{ll}
[0,0]^\transpose & \textrm{ if } \delta_i(\v{x}) = \textrm{ false }\\
\nonumber{} \omega_i [\sin{\pi\rho_i\frac{x_i}{u_i-l_i}}, \cos{\pi\rho_i\frac{x_i}{u_i-l_i}}]^\transpose & \textrm{ otherwise.}\end{array}\right..
\end{eqnarray}

Although our formal arguments do not rely on this, Proposition \ref{prop:dbr_pseudometric} in the appendix shows that $d\br_i$ is a pseudometric. 
This pseudometric is defined by two parameters: $\omega_i \in [0,1]$ and $\rho_i \in [0,1]$. We firstly define 
\begin{equation}\label{eq:gamma}
\omega_i = \prod_{j \in \text{anc}_i \cup \{i\}} \gamma_j, 
\end{equation}
where $\gamma_j \in [0,1]$. This encodes the intuitive notion that differences on lower levels of the hierarchy count less than differences in their ancestors.

Also note that, as desired, if $i$ is inactive for both $\v{x}$ and $\v{x}'$, $d\br_i$ specifies that $g(\v{x})$ and $g(\v{x}')$ should not differ owing to differences between $x_i$ and $x_i'$. Secondly, if $i$ is active for both $\v{x}$ and $\v{x}'$, the difference between $g(\v{x})$ and $g(\v{x}')$ due to $x_i$ and $x_i'$ increases monotonically with increasing $\left|x_i-x_i'\right|$. Parameter $\rho_i$ controls whether differing in the activity of $i$ contributes more or less to the distance than differing in $x_i$ should $i$ be active. If $\rho = \nicefrac{1}{3}$, and if $i$ is inactive for exactly one of $\v{x}$ and $\v{x}'$, $g(\v{x})$ and $g(\v{x}')$ are as different as is possible due to dimension $i$; that is, $g(\v{x})$ and $g(\v{x}')$ are exactly as different in that case as if $x_i=l_i$ and $x_i'=u_i$. For $\rho>\nicefrac{1}{3}$, $i$ being active for both $\v{x}$ and $\v{x}'$ means that $g(\v{x})$ and $g(\v{x}')$ could potentially be more different than if
$i$ was active in only one of them. For $\rho<\nicefrac{1}{3}$, the converse is true.\footnote{Note that $\v{x}$ and $\v{x}'$ must differ in at least one ancestor dimension of $i$ in order for $\delta_i(\v{x}) \neq \delta_i(\v{x}')$ to hold, such that in the final kernel combining kernels $k_i$ due to each dimension $i$, differences in the activity of dimension $i$ are penalized both in kernel $k_i$ and in the distance for the kernel of the ancestor dimension causing the difference in $i$'s activity.}

We now show that $d\br_i$ and $f\br_i$ can be plugged into a positive semi-definite kernel over Euclidean space to define a valid kernel over space $\sX$.

\begin{prop}
Let $\kappa$ be a positive semi-definite covariance function over Euclidean space.
Then, $k_i\colon \sX \times \sX\to \reals^+$, defined by 
%\[k_i(\v{x},\v{x}') = \kappa( d_{\text{E}}(f_i(\v{x}), f_i(\v{x}')) )\]
\[k_i(\v{x},\v{x}') = \kappa\bigl( d\br_i(\v{x}, \v{x}') \bigr)\]
is a positive semi-definite covariance function over input space $\sX$. 
\label{prop:cont_psd}
\begin{proof}
Due to Proposition \ref{prop:psd_if_isometry}, we only need to show that, for any two inputs $\v{x},\v{x}' \in \sX$, the isometry condition $d_{\text{E}}\bigl(f_i\br(\v{x}),f_i\br(\v{x}')\bigr) = d\br_i(\v{x},\v{x}')$ holds.

We use the abbreviation $\alpha = \pi\rho_i\frac{x_i}{u_i-l_i}$ and $\alpha' = \pi\rho_i\frac{x'_i}{u_i-l_i}$ and consider the following three possible cases of dimension $i$ being active or inactive in $\v{x}$ and $\v{x}'$.

~\\\noindent{}{Case 1}: $\delta_i(\v{x}) = \delta_i(\v{x}') = \textrm{false}$.
In this case, we trivially have 
\[d_{\text{E}}(f_i\br(\v{x}),f_i\br(\v{x}')) = d_{\text{E}}([0,0]^\transpose, [0,0]^\transpose) = 0 = d\br_i(\v{x},\v{x}').\]

~\\\noindent{}{Case 2}: $\delta_i(\v{x}) \neq \delta_i(\v{x}')$. In this case, we have
\[d_{\text{E}}(f_i\br(\v{x}),f_i\br(\v{x}')) = d_{\text{E}}([\sin{\alpha}, \cos{\alpha}]^\transpose, [0,0]^\transpose) = \sqrt{\omega_i^2 (\sin^2{\alpha} + \cos^2{\alpha})} = \omega_i = d\br_i(\v{x},\v{x}'),\]
and symmetrically for $d_{\text{E}}([0,0]^\transpose, [\sin{\alpha}, \cos{\alpha}]^\transpose)$.

~\\\noindent{}{Case 3}: $\delta_i(\v{x}) = \delta_i(\v{x}') = \textrm{true}$. We have:
\begin{eqnarray}
\nonumber{}d_{\text{E}}(f_i\br(\v{x}),f_i\br(\v{x}')) & = & d_{\text{E}}(\omega_i [\sin{\alpha}, \cos{\alpha}]^\transpose, \omega_i [\sin{\alpha'}, \cos{\alpha'}]^\transpose)\\ 
\nonumber{}& = & \omega_i \sqrt{(\sin{\alpha}-\sin{\alpha'})^2+ (\cos{\alpha}-\cos{\alpha'})^2}\\
\nonumber{}& = & \omega_i \sqrt{\sin^2{\alpha} -2 \sin{\alpha}\sin{\alpha'} + \sin^2{\alpha'}  + \cos^2{\alpha} -2 \cos{\alpha}\cos{\alpha'} + \cos^2{\alpha'} }\\
\nonumber{}& = & \omega_i \sqrt{(\sin^2{\alpha}+\cos^2{\alpha})   +  (\sin^2{\alpha'}+\cos^2{\alpha'})   -2 (\sin{\alpha}\sin{\alpha'} + \cos{\alpha}\cos{\alpha'})}\\
\label{eqn:simplified}& = & \omega_i \sqrt{ 1+1-2 \cos(\alpha-\alpha')}\\
\nonumber& = & \omega_i \sqrt{2} \sqrt{1 - \cos(\pi\rho_i \frac{x_i-x_i'}{u_i-l_i})} = d\br_i(\v{x}, \v{x}'),
\end{eqnarray}
where (\ref{eqn:simplified}) follows from the previous line by using the identity 
\[\cos{(a-b)} = \cos{a}\cos{b} + \sin{a}\sin{b}.\]
\end{proof}
\end{prop}

\section{Categorical Dimensions}
%%%%%%%%%%%%%%%%%%%%%%%%%%%%%%%%%%%%%%%%%%%%%%%%%%%%%%%%%%%%%%%%%%%%%%%%%%%%%%%%%%%%%%%%%%%%%%%%%%%%%%%%%%%%%%%%%%%%%%%%%%%%%

Now let's define $f\cat_i$ and $d\cat_i$ for the case that the input $\sX_i=\{v_{i,1}, \dots, v_{i,m_i}\}$ is categorical with $m_i$ possible values. 
Proceeding as above, we define a pseudometric $d\cat_i$ on $\sX$ and an isometry from $(\sX, d\cat_i)$ to $(\reals^{m_i},d_{\text{E}}^{m_i})$, and show that we can combine these
with a kernel over Euclidean space to construct a valid kernel over space $\sX$. 

\begin{eqnarray}
\nonumber{}d\cat_i(\v{x}, \v{x}') & = & \left\{
\begin{array}{ll}
\nonumber{} 0 & \textrm{ if } \delta_i(\v{x}) = \delta_i(\v{x}') = \textrm{false}\\
\nonumber{} \omega_i & \textrm{ if } \delta_i(\v{x}) \neq \delta_i(\v{x}')\\
\nonumber{} \omega_i \frac{\sqrt{2} \rho}
{1+(m_i-1)(1-\rho)^2}
 \indicator_{x_i \neq x_i'} 
& \textrm{ if } \delta_i(\v{x}) = \delta_i(\v{x}') = \textrm{true}.
\end{array}
\right.
\end{eqnarray}

\begin{eqnarray}
\nonumber{}f\cat_i(\v{x}) & = & \left\{\begin{array}{ll}
\v{0} \in \reals^{m_i} & \textrm{ if } \delta_i(\v{x}) = \textrm{ false }\\
\nonumber{} \omega_i\,\frac{\v{e_j}+(1-\rho)\sum_{l\neq j} \v{e_l}}
{\sqrt{1+(m_i-1)(1-\rho)^2}}
 & \textrm{ if } \delta_i(\v{x}) = \textrm{ true and } x_i = v_{i,j},
\end{array}\right.
\end{eqnarray}
\noindent{}where $\v{e_j} \in \reals^{m_i}$ is the $j$th unit vector: zero in all dimensions except $j$, where it is $1$. Note that
\begin{equation}
 \sqrt{1+(m_i-1)(1-\rho)^2} = \biggl\|\v{e_j}+(1-\rho)\sum_{l\neq j} \v{e_l}\biggr\|.
\end{equation}

\noindent{}Again, although our analysis does not require it, we prove in Proposition \ref{prop:dbr_pseudometric_cat} (see appendix) that $d\cat_i$ is a pseudometric. Our pseudometric is again defined by two hyperparameters. Firstly, $\omega_i\in[0,1]$ is exactly as defined in \eqref{eq:gamma}, and similarly allows higher-level inputs to attain greater importance. Similarly, $\rho_i\in[0,1]$ allows control of to what extent differing in the activity of $i$ affects the distance relative to the influence of differing in $x_i$ should $i$ be active. In particular, for
\begin{equation}
 \rho_i^\ast = 
\frac{\sqrt{2}-2+2m_i-\sqrt{6-4\sqrt{2}+4(\sqrt{2}-1)m_i}}
{2(m_i-1)},
\end{equation}
$\rho_i<\rho_i^\ast$ implies that differing in the activity of $i$ is more significant, whereas $\rho_i>\rho_i^\ast$ implies the converse. The special case $\rho_i = 0$ dictates that differing in $x_i$ has no influence on the distance; $\rho_i=1$ assigns maximal importance to differing in $x_i$. 

\begin{prop}
Let $\kappa$ be a positive semi-definite covariance function over Euclidean space.
Then, $k_i\colon \sX \times \sX\to \reals^+$, defined by 
%\[k_i(\v{x},\v{x}') = \kappa( d_{\text{E}}(f_i(\v{x}), f_i(\v{x}')) )\]
\[k_i(\v{x},\v{x}') = \kappa\bigl( d\cat_i(\v{x}, \v{x}') \bigr)\]
is a positive semi-definite covariance function over input space $\sX$. 
\label{prop:cat_psd}
\begin{proof}
We proceed as in the proof of Proposition \ref{prop:cont_psd} to show that, for any two inputs $\v{x},\v{x}' \in \sX$, the isometry condition $d_{\text{E}}^{m_i}(f\cat_i(\v{x}),f\cat_i(\v{x}')) = d\cat_i(\v{x},\v{x}')$ holds.

~\\\noindent{}{Case 1}: $\delta_i(\v{x}) = \delta_i(\v{x}') = \textrm{false}$.
In this case, we trivially have 
\[d_{\text{E}}^{m_i}(f_i\br(\v{x}),f_i\br(\v{x}')) = d_{\text{E}}^{m_i}(\v{0}, \v{0}) = 0 = d_i\br(\v{x},\v{x}').\]

~\\\noindent{}{Case 2}: $\delta_i(\v{x}) \neq \delta_i(\v{x}')$. In this case, we have
\[d_{\text{E}}^{m_i}(f\cat_i(\v{x}),f\cat_i(\v{x}')) = 
d_{\text{E}}^{m_i}\biggl(\omega_i\,\frac{\v{e_j}+(1-\rho)\sum_{l\neq j} \v{e_l}}
{\|\v{e_j}+(1-\rho)\sum_{l\neq j} \v{e_l}\|}, \v{0}\biggr) 
= \omega_i = d_i(\v{x},\v{x}'),\]
and symmetrically for $d_{\text{E}}\biggl(\v{0}, \omega_i\,\frac{\v{e_j}+(1-\rho)\sum_{l\neq j} \v{e_l}}
{\|\v{e_j}+(1-\rho)\sum_{l\neq j} \v{e_l}\|}\biggr)$.

~\\\noindent{}{Case 3}: $\delta_i(\v{x}) = \delta_i(\v{x}') = \textrm{true}$. 
If $x_i=x_i'=v_{i,j}$, we have 
\begin{eqnarray}
\nonumber{}d_{\text{E}}^{m_i}(f\cat_i(\v{x}),f\cat_i(\v{x}')) & = & d_{\text{E}}^{m_i}\bigl(
f\cat_i(\v{x}),f\cat_i(\v{x})
\bigr) = 0 = d\cat_i(\v{x}, \v{x}').
\end{eqnarray}

\noindent{}If $x_i=v_{i,j} \neq v_{i,j'} = x_i'$, we have 
\begin{align} 
\nonumber{}d_{\text{E}}(f\cat_i(\v{x}),f\cat_i(\v{x}')) & = 
d_{\text{E}}^{m_i}\biggl(
\omega_i\,\frac{\v{e_j}+(1-\rho)\sum_{l\neq j} \v{e_l}}
{\sqrt{1+(m_i-1)(1-\rho)^2}},\,
\omega_i\,\frac{\v{e_j'}+(1-\rho)\sum_{l\neq j'} \v{e_l}}
{\sqrt{1+(m_i-1)(1-\rho)^2}}
\biggr) \\
& = \omega_i \frac{\sqrt{\bigl(1-(1-\rho)\bigr)^2 + \bigl(1-(1-\rho)\bigr)^2}}
{1+(m_i-1)(1-\rho)^2} \nonumber\\
& = \omega_i \frac{\sqrt{2} \rho}
{1+(m_i-1)(1-\rho)^2} \nonumber\\
& = d\cat_i(\v{x}, \v{x}').
\end{align}
\end{proof}
\end{prop}

\appendix

\section{Proof of pseudometric properties}

\begin{prop}
  $d\br_i$ is a pseudometric on $\sX$. \label{prop:dbr_pseudometric}
\begin{proof}
The non-negativity and symmetry of $d\br_i$ are trivially proven. To prove the triangle inequality, consider $\v{x}, \v{x}', \v{x}'' \in \sX$. 

~\\\noindent{}{Case 1}: $\delta_i(\v{x}) = \delta_i(\v{x}') = \textrm{false}$, such that $d\br_i(\v{x},\v{x}') = 0$. Here, from non-negativity, clearly $d\br_i(\v{x},\v{x}') = 0 \leq d\br_i(\v{x},\v{x}'') + d\br_i(\v{x}',\v{x}'')$.

~\\\noindent{}{Case 2}: $\delta_i(\v{x}) \neq \delta_i(\v{x}')$, such that such that  $d\br_i(\v{x},\v{x}') = \omega_i$.  Without loss of generality, assume $\delta_i(\v{x}) = \text{true}$, $\delta_i(\v{x}') = \text{false}$ and $\delta_i(\v{x}'') = \text{true}$. 
\begin{align}
d\br_i(\v{x},\v{x}'') + d\br_i(\v{x}',\v{x}'') = d\br_i(\v{x},\v{x}'')  + \omega_i
\end{align}
Hence $d\br_i(\v{x},\v{x}'') + d\br_i(\v{x}',\v{x}'') \geq \omega_i = d\br_i(\v{x},\v{x}')$ by non-negativity.

~\\\noindent{}{Case 3}: $\delta_i(\v{x}) = \delta_i(\v{x}')=\textrm{true}$, such that  $d\br_i(\v{x},\v{x}') = \omega_i \sqrt{2} \sqrt{1 - \cos(\pi\rho_i \frac{x_i-x_i'}{u_i-l_i})}$.  If  $\delta_i(\v{x}'') = \text{false}$,
\begin{align}
d\br_i(\v{x},\v{x}'') + d\br_i(\v{x}',\v{x}'') = 2 \omega_i \geq \omega_i \sqrt{2} \sqrt{1 - \cos(\pi\rho_i \frac{x_i-x_i'}{u_i-l_i})} = d\br_i(\v{x},\v{x}').
\end{align} 
If  $\delta_i(\v{x}'') = \text{true}$, consider the `worst' possible case in which, without loss of generality, $x_i=l_i$ and $x'_i=u_i$, such that $d\br_i(\v{x},\v{x}')=2 \omega_i^2$.  We define the abbreviation $\beta'' = \frac{x''_i-l_i}{u_i-l_i}$, giving
\begin{align}
\bigl(d\br_i(\v{x},\v{x}'') + d\br_i(\v{x}',\v{x}'')\bigr)^2
& = 2\omega_i^2 \Bigl(\sqrt{1 - \cos (\pi\rho_i \beta'')} + \sqrt{1 - \cos \bigl(\pi\rho_i (1-\beta'')\bigr)}\Bigr)^2\nonumber\\
&=2\omega_i^2\biggl(2 - \cos (\pi\rho_i \beta'') - \cos \bigl(\pi\rho_i (1-\beta'')\bigr)
\nonumber\\
&\qquad\qquad+2\sqrt{\Bigl(1 - \cos (\pi\rho_i \beta'')\Bigr)\Bigl(1 - \cos \bigl(\pi\rho_i (1-\beta'')\bigr)\Bigr)}\biggr)\nonumber\\
&=2\omega_i^2\biggl(2 +2\sqrt{1 + \cos (\pi\rho_i \beta'')\cos \bigl(\pi\rho_i (1-\beta'')\bigr)}\biggr)\nonumber\\
&=4 \omega_i^2 \bigl(1 + \left|\sin \pi\rho_i \beta'' \right|\bigr)\nonumber\\
&\geq 4 \omega_i^2 = d\br_i(\v{x},\v{x}')^2.
\end{align}
Hence, from non-negativity, we have $d\br_i(\v{x},\v{x}'') + d\br_i(\v{x}',\v{x}'')\geq d\br_i(\v{x},\v{x}')$.
\end{proof}
\end{prop}

\begin{prop}
 $d\cat_i$ is a pseudometric on $\sX$.\label{prop:dbr_pseudometric_cat}
\begin{proof}
The non-negativity and symmetry of $d\cat_i$ are trivially proven. To prove the triangle inequality, consider $\v{x}, \v{x}', \v{x}'' \in \sX$. 

~\\\noindent{}{Case 1}: $\delta_i(\v{x}) = \delta_i(\v{x}') = \textrm{false}$, such that $d\cat_i(\v{x},\v{x}') = 0$. Here, from non-negativity, clearly $d\cat_i(\v{x},\v{x}') = 0 \leq d\cat_i(\v{x},\v{x}'') + d\cat_i(\v{x}',\v{x}'')$.

~\\\noindent{}{Case 2}: $\delta_i(\v{x}) \neq \delta_i(\v{x}')$, such that such that  $d\cat_i(\v{x},\v{x}') = \omega_i$.  Without loss of generality, assume $\delta_i(\v{x}) = \text{true}$, $\delta_i(\v{x}') = \text{false}$ and $\delta_i(\v{x}'') = \text{true}$. 
\begin{align}
d\cat_i(\v{x},\v{x}'') + d\cat_i(\v{x}',\v{x}'') = d\cat_i(\v{x},\v{x}'')  + \omega_i
\end{align}
Hence $d\cat_i(\v{x},\v{x}'') + d\cat_i(\v{x}',\v{x}'') \geq \omega_i = d\cat_i(\v{x},\v{x}')$ by non-negativity.

~\\\noindent{}{Case 3}: $\delta_i(\v{x}) = \delta_i(\v{x}')=\textrm{true}$, such that  $d\cat_i(\v{x},\v{x}') =
\omega_i \frac{\sqrt{2} \rho}
{1+(m_i-1)(1-\rho)^2}
 \indicator_{x_i \neq x_i'} $.  
If  $\delta_i(\v{x}'') = \text{false}$,
\begin{align}
d\cat_i(\v{x},\v{x}'') + d\cat_i(\v{x}',\v{x}'') = 
2 \omega_i 
\geq 
\omega_i \frac{\sqrt{2} \rho}
{1+(m_i-1)(1-\rho)^2}
 \indicator_{x_i \neq x_i'}  
= d\cat_i(\v{x},\v{x}').
\end{align} 
If  $\delta_i(\v{x}'') = \text{true}$, 
\begin{align}
d\cat_i(\v{x},\v{x}'') + d\cat_i(\v{x}',\v{x}'')
& = \omega_i \frac{\sqrt{2} \rho}
{1+(m_i-1)(1-\rho)^2}
(
 \indicator_{x_i \neq x_i''} 
+
 \indicator_{x_i' \neq x_i''} 
)
\nonumber\\
& \geq 
\omega_i \frac{\sqrt{2} \rho}
{1+(m_i-1)(1-\rho)^2}
 \indicator_{x_i \neq x_i'} = d\cat_i(\v{x},\v{x}').
\end{align}
\end{proof}
\end{prop}

%\bibliographystyle{theapa}
%\renewcommand{\baselinestretch}{0.97}
%\footnotesize{\bibliography{abbrev,frankbib}}

\end{document}